\documentclass{article}

 \usepackage[preprint, nonatbib]{neurips_2026}

\usepackage[utf8]{inputenc} 
\usepackage[T1]{fontenc}    
\usepackage{hyperref}       
\usepackage{url}            
\usepackage{booktabs}       
\usepackage{amsfonts}       
\usepackage{nicefrac}       
\usepackage{microtype}      
\usepackage[table]{xcolor}         
\usepackage{amsmath}

\usepackage{multirow}
\usepackage{graphicx}
\usepackage{subcaption}

\usepackage{amsthm}

\usepackage{pifont}
\newcommand{\cmark}{\ding{51}}
\newcommand{\xmark}{\ding{55}}

\title{One-Step Distillation of Discrete Diffusion Image Generators via Fixed-Point Iteration}

\author{%
  Chaoyang Wang \quad Yunhai Tong \\
  Peking University\\
  \texttt{cywang@stu.pku.edu.cn} \\
}

\begin{document}

\maketitle

\begin{figure}[ht]
	\centering
	\includegraphics[width=1.0\linewidth]{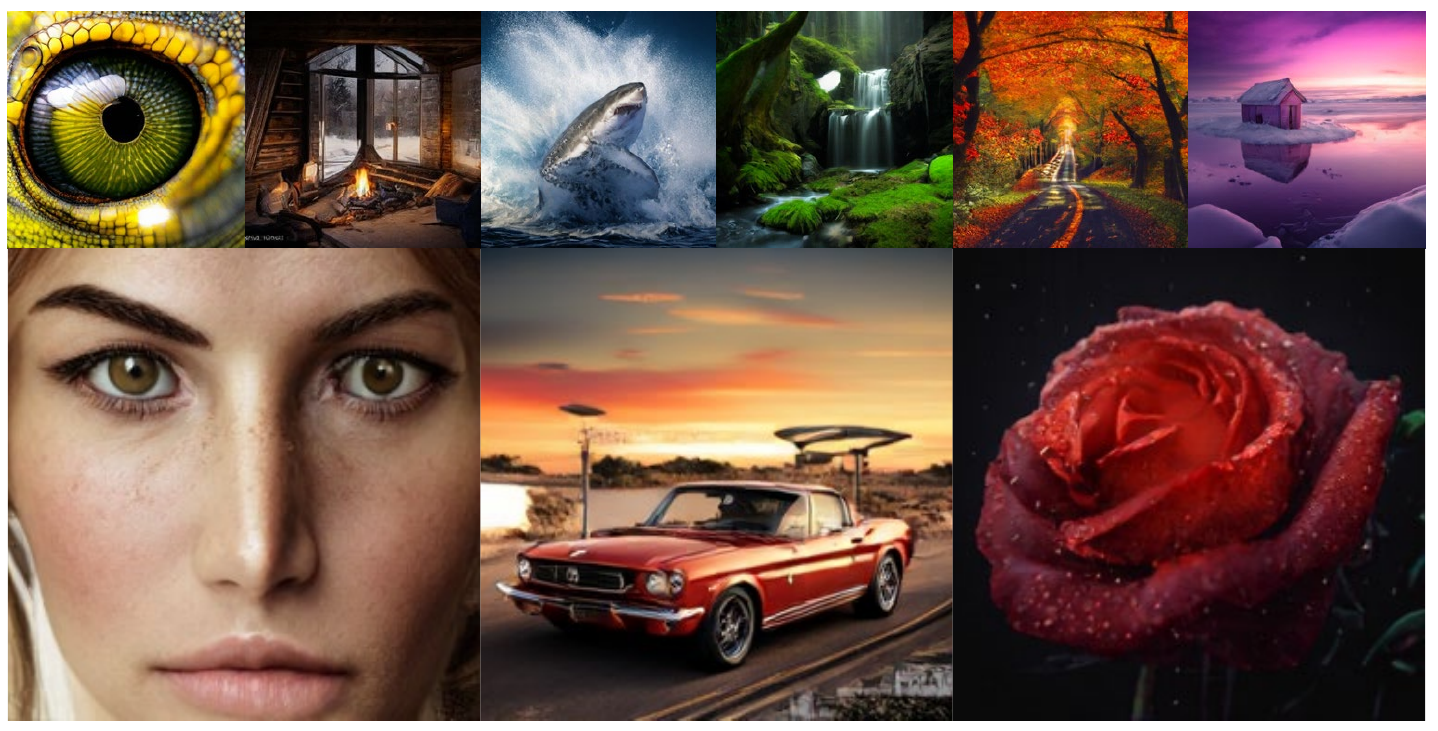}
	\caption{Text-to-image samples from MaskGen-L distilled with our Fixed-Point Distillation (FPD) framework, where each image is synthesized in a single forward pass.}
    \label{fig:teaser}
\end{figure}

\begin{abstract}

Discrete diffusion models excel at visual synthesis but rely on slow, iterative decoding. Existing single-step distillation methods attempt to bypass this
bottleneck, either by training auxiliary score networks that effectively double
compute, or by introducing specialized parameterizations and multi-stage
pipelines that fragment optimization.
In this paper, we introduce Fixed-Point Distillation (FPD), an end-to-end framework that constructs local correction targets by partially corrupting the student's one-step draft and refining it with a single teacher step. To compute the training objective in a semantically meaningful space, we lift discrete tokens into continuous features and apply a multi-bandwidth drift loss that iteratively accumulates these corrections.
To backpropagate through the discrete bottleneck, we employ a straight-through estimator that feeds exact hard-sampled tokens to the teacher and decoder during the forward pass, ensuring that training and inference operate on the same codebook manifold, while routing continuous gradients back to the student logits. This fully differentiable pathway additionally accommodates an optional unconditional adversarial objective to enhance perceptual realism.
Evaluations on both class- and text-conditional generation validate the
effectiveness of our framework. FPD achieves competitive visual fidelity and
structural alignment within a single inference step, narrowing the gap to
multi-step teachers while outperforming existing discrete distillation
baselines.

\end{abstract}

\section{Introduction}
\label{sec:intro}

Generative models based on iterative refinement have emerged as the leading paradigm for high-fidelity visual synthesis. By breaking down the generative process into a sequence of transitional steps, these models accurately capture complex data distributions, driving rapid progress across text-to-image generation~\cite{ramesh2021zero,ramesh2022hierarchical,rombach2022high,podell2023sdxl,saharia2022photorealistic,nichol2021glide}, video synthesis~\cite{singer2022make,ho2022video,yan2021videogpt,hong2022cogvideo}, image editing~\cite{brooks2023instructpix2pix,meng2021sdedit}, 3D asset creation~\cite{poole2022dreamfusion}, and world modeling~\cite{agarwal2025cosmos,bruce2024genie}. 
Among them, discrete diffusion models that operate on quantized token sequences have attracted growing attention for their natural compatibility with language-based architectures and scalable training. However, these methods typically share a fundamental computational bottleneck:
the requirement for tens or hundreds of sequential network evaluations per sample. This high inference latency severely restricts their utility in latency-sensitive environments, motivating an intensive search for efficient acceleration methods.

An effective remedy is \emph{distillation}, which trains a student model to reproduce the output of a pretrained teacher in far fewer steps. In the continuous domain, the literature has largely converged around three successful paradigms: trajectory imitation~\cite{salimans2022progressive,song2023consistency,luo2023latent,lu2024simplifying,geng2025mean}, distribution matching~\cite{yin2024one,yin2024improved,luo2023diff,zhou2024score}, and adversarial post-training~\cite{sauer2024adversarial,sauer2024fast}. These approaches now routinely yield robust one- to four-step generators. In contrast, the distillation of discrete masked diffusion models has advanced far more slowly. Existing discrete distillers generally follow one of two suboptimal paths. They either adapt the continuous score-distillation recipe by relying on an online auxiliary network~\cite{zhu2025di,zhu2025soft}, which drastically inflates training compute and memory, or introduce correlation-aware objectives through bespoke designs --- a mixture-model student parameterization~\cite{hayakawa2024distillation} or multi-stage coupling rectification~\cite{yoo2025redi} --- that bypass the auxiliary network but inflate per-step computation or fragment training into iterative offline stages. Neither route yields a single-step discrete distiller that is simultaneously efficient to train and universally applicable across different masking schedules. To overcome these limitations, we step away from artificial trajectory tracking and auxiliary surrogate score networks altogether, proposing a fundamentally different distillation paradigm.

In this work, we propose a principled, end-to-end distillation framework that fundamentally casts single-step generation as a fixed-point matching problem. \textbf{Our core motivation is to repurpose the frozen teacher as a state-dependent local refinement operator, rather than regressing toward pre-collected multi-step teacher rollouts.} Given a pretrained multi-step teacher, our method trains the student exclusively on its own generated drafts. Specifically, starting from a nearly fully masked token sequence with random codebook entries, the student produces a complete draft in a single forward pass. This draft is then partially re-masked and refined by the teacher through a single denoising correction, yielding a target tightly coupled to the student's current output. At convergence, the student's output distribution becomes invariant under this corruption-and-refinement cycle, meaning that single-step generation constitutes a fixed point of the distillation process.
To compute the training objective in a meaningful metric space, we lift both the student's draft and the teacher's refinement into a shared continuous feature space via a frozen encoder. Since discrete tokens admit no well-defined distance, this lifted space allows us to apply a multi-bandwidth drift loss that provides stable gradient signals without auxiliary score networks. As training progresses, these local corrections accumulate, gradually steering the student's generation toward the data manifold.

However, backpropagating the drift loss through the discrete sampling step is nontrivial. Conventional continuous relaxations, such as soft probability mixtures over codebook embeddings, produce off-manifold inputs that distort both the teacher's refinement and the decoder's reconstruction, introducing persistent bias in the training target. To resolve this, we employ a straight-through estimator (STE) that feeds exact hard-sampled tokens in the forward pass, ensuring that the decoder always receives valid discrete inputs, while routing continuous gradients from the drift loss back to the student logits during the backward pass. 
This ensures that training and inference operate on identical discrete
inputs, eliminating the train-test mismatch inherent in soft relaxations. 
Moreover, the fully differentiable pathway from pixel space to logits additionally accommodates an optional unconditional adversarial loss on the decoded student images, which enhances perceptual realism without sacrificing the semantic alignment enforced by the drift objective.

Our contributions are summarized as follows: \textbf{1)} We propose Fixed-Point Distillation, a framework that formulates single-step discrete distillation as a fixed-point matching problem, bypassing complex trajectory tracking, pre-collected multi-step teacher samples, auxiliary score networks, or fragmented training pipelines. \textbf{2)} We introduce a lifted drift objective paired with a straight-through estimator, enabling fully end-to-end training through the discrete bottleneck with consistent training and inference behavior. \textbf{3)} Empirical experiments and ablations verify the effectiveness, whilst demonstrating competitive performance against existing discrete distillation methods.

\section{Related Work}
\label{sec:related_work}

\subsection{Discrete Diffusion Models}

Diffusion models learn to generate data by progressively corrupting it with a forward noising process and training a network to reverse the corruption. The continuous-domain lineage, spanning from early denoising models~\cite{ho2020denoising,song2020score} to modern flow matching architectures, has produced the dominant large-scale visual generators~\cite{esser2024scaling,rombach2022high,podell2023sdxl}. A parallel discrete branch works on quantized token sequences and forms the primary focus of this work. In the language domain, masked diffusion and score-entropy formulations~\cite{sahoo2024simple,lou2023discrete} implement absorbing- and uniform-state corruption processes, demonstrating generation capabilities that rival those of traditional autoregressive (AR) decoders at comparable scales. 
For visual synthesis, a growing line of work extends discrete diffusion to
image generation~\cite{austin2021structured,campbell2022continuous,shi2024simplified,bond2022unleashing,gu2022vector,hu2022unified,sun2022score}.
Among them, masked generative transformers such as
MaskGIT~\cite{chang2022maskgit} and
MaskGen~\cite{kim2025democratizing} iteratively predict and reveal
tokens over a fixed schedule, differing primarily in their tokenization
strategies and conditioning mechanisms. Regardless of these design choices, discrete diffusion models typically require tens to hundreds of sequential forward
passes per sample, severely limiting their deployment in latency-sensitive
settings.

\subsection{Discrete Diffusion Distillation}

While continuous-domain distillation has matured rapidly through trajectory imitation~\cite{salimans2022progressive,song2023consistency,lu2024simplifying,geng2025mean}, distribution matching~\cite{yin2024one,yin2024improved}, and adversarial frameworks~\cite{sauer2024adversarial,sauer2024fast}, the literature on discrete distillation remains relatively sparse. Progress has largely relied on porting continuous primitives into categorical token spaces. This introduces severe structural challenges: the non-differentiability of token sampling, the lack of deterministic latent trajectories under masked corruption, and the factorization error inherent in single-step predictions.

To navigate these challenges, recent methods in both visual and language domains explore diverse adaptations. 
One prominent direction adapts distribution matching to discrete outputs by minimizing token-level divergences. 
Di[M]O performs on-policy token-level distribution matching with an auxiliary model initialized from the teacher, which approximates the student's intermediate conditional distributions and provides gradients directly on generator logits, avoiding backpropagation through discrete sampling~\cite{zhu2025di}. 
Soft-Di[M]O further relaxes discrete outputs into soft embeddings, enabling end-to-end differentiable refinements such as adversarial fine-tuning~\cite{zhu2025soft}. 
Alternative approaches mitigate product-factorization errors by parameterizing student transitions as mixtures of product models to capture dimensional correlations~\cite{hayakawa2024distillation}, or by iteratively rectifying source--target couplings to reduce conditional total correlation in discrete flows~\cite{yoo2025redi}. 
In the language domain, SDTT distills masked discrete diffusion language models by matching student predictions to teacher-generated multi-step denoising targets~\cite{deschenaux2024beyond}, while D-MMD generalizes continuous moment-matching distillation to discrete/categorical diffusion models~\cite{hoogeboom2026beyond}.

Crucially, existing discrete distillation methods typically incur substantial computational overhead, either through online learnable auxiliary networks that effectively double training cost, or through multi-stage pipelines that fragment optimization into separate offline phases. Our framework sidesteps both issues by formulating distillation as a fixed-point matching problem, replacing trajectory-level imitation with local corrections derived from the student's own drafts, requiring no additional trainable components beyond the student and remaining agnostic to the teacher's masking schedule.

\section{Method}
\label{sec:method}

We first review the discrete diffusion framework in
Sec.~\ref{subsec:preliminary}, then present our fixed-point formulation that
constructs training targets from the student's own re-masked drafts in
Sec.~\ref{subsec:fpd}. The resulting residual is computed via a
multi-bandwidth drift loss in a lifted continuous feature space, as detailed
in Sec.~\ref{subsec:drift}. Finally, Sec.~\ref{subsec:stein} describes the
straight-through estimator that bridges the discrete bottleneck and enables
an optional adversarial objective.

\begin{figure}[t!]
	\centering
	\includegraphics[width=1.0\linewidth]{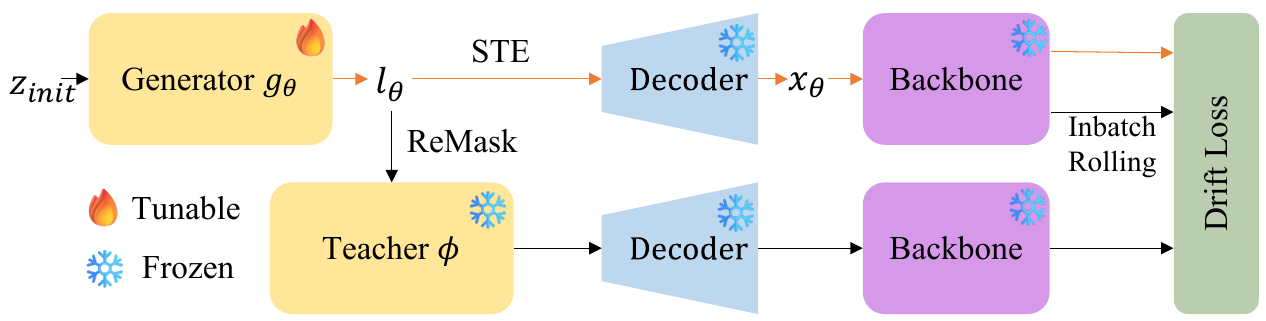}
	\caption{\textbf{Overview of the proposed distillation framework.} The student model processes a masked initialization $z_{\text{init}}$ to output logits $\ell_\theta$. Via a Straight-Through Estimator (STE), these discrete outputs are decoded into a continuous image $x_\theta$ and mapped to a feature space by a frozen backbone $\Phi$. Concurrently, the student's token draft is partially re-masked ($\mathcal{M}_r$) and refined by the frozen teacher. The decoded teacher features serve as positive targets, while cyclically shifted in-batch student features act as negatives to compute the drift loss $\mathcal{L}_{\text{drift}}$. An auxiliary GAN loss $\mathcal{L}_{\text{gan}}$ is applied to $x_\theta$ for enhanced perceptual quality. Black lines denote forward computation paths, while red lines highlight the differentiable gradient flow that safely bypasses the discrete bottleneck. The backbone and decoder share the same weights.}
    \label{fig:method}
\end{figure}

\subsection{Preliminaries: Discrete Diffusion for Image Generation}
\label{subsec:preliminary}

Let $x\in\mathbb{R}^{3\times H\times W}$ be an image and let $(\mathcal{E},\mathcal{D})$ denote a frozen VQ encoder-decoder pair with codebook $E\in\mathbb{R}^{K\times d}$, where $K$ is the codebook size and $d$ the embedding dimension. The encoder maps an image to a length-$L$ sequence of discrete codes $z=(z_1,\ldots,z_L)\in\mathcal{V}^L$ with $\mathcal{V}=\{1,\ldots,K\}$, and the decoder reconstructs an image via $\mathcal{D}(E[z])$. We write $E[z]\in\mathbb{R}^{L\times d}$ for the row-wise codebook lookup. Text conditioning $c$ is produced by a separate frozen text encoder.

\noindent \textbf{Forward masking process.} Discrete diffusion augments the token vocabulary with a special mask symbol $[\textsc{m}]\notin\mathcal{V}$, yielding $\bar{\mathcal{V}}=\mathcal{V}\cup\{[\textsc{m}]\}$. For a continuous timestep $t\in[0,1]$ and a monotonically increasing noise schedule $\gamma:[0,1]\to[0,1]$, each position is independently replaced with $[\textsc{m}]$ with probability $\gamma(t)$:
\begin{equation}
q(z_t\mid z)=\prod_{i=1}^{L}\big[(1-\gamma(t))\,\delta_{z_{t,i},z_i}+\gamma(t)\,\delta_{z_{t,i},[\textsc{m}]}\big].
\end{equation}

\noindent \textbf{Reverse unmasking process.} A teacher network $f_\phi:\bar{\mathcal{V}}^L\times\mathcal{C}\times[0,1]\to\mathbb{R}^{L\times K}$ outputs per-position logits $\ell=f_\phi(z_t,c,t)$, which define a categorical distribution $p_\phi(z_{0,i}=v\mid z_t,c,t)\propto\exp(\ell_{i,v})$, where $\mathcal{C}$ denotes the text conditioning space. The training objective minimizes the cross-entropy loss on the masked positions,
\begin{equation}
\mathcal{L}_\phi=\mathbb{E}_{t,\,z\sim p_\text{data},\,z_t\sim q(\cdot\mid z)}\bigg[-\!\!\sum_{i:\,z_{t,i}=[\textsc{m}]}\!\!\log p_\phi(z_i\mid z_t,c,t)\bigg].
\end{equation}
During inference, generation proceeds iteratively from a fully masked sequence $z_T=([\textsc{m}],\ldots,[\textsc{m}])$. At each of the $T$ steps, the model predicts all masked positions, retains a top-confidence subset, and re-masks the remainder according to $\gamma$. We denote a single such teacher refinement step at mask ratio $r$ by $\mathcal{T}_\phi^{r}$. After $T$ steps, the final tokens $\hat{z}$ are decoded to pixels via $\hat{x}=\mathcal{D}(E[\hat{z}];\,c)$.

\noindent \textbf{Distillation.} To eliminate this iterative overhead, our distillation framework trains a single-step student $g_\theta$ that directly maps a masked initialization to a high-quality sample, bypassing the teacher's multi-step inference process. The following section details our approach.

\subsection{Fixed-Point Iteration via Local Correction}
\label{subsec:fpd}

The teacher's reverse process is a sequence of refinements $\mathcal{T}_\phi^{r_1}\!\circ\!\mathcal{T}_\phi^{r_2}\!\circ\cdots$ that progressively reduce the masking level. A correctly distilled single-step student $g_\theta$ should produce, in one forward pass, a sample $\hat{z}$ that is already a \emph{fixed point} of the teacher's refinement operator: re-masking $\hat{z}$ at any level $r$ and applying $\mathcal{T}_\phi^{r}$ should leave its distribution invariant. Writing $p_\theta$ for the student's marginal, this requirement can be stated as
\begin{equation}\label{eq:fp}
p_\theta=\big(\mathcal{T}_\phi^{r}\circ\mathcal{M}_r\big)_{\#}\,p_\theta,\qquad \forall\,r\in(0,1),
\end{equation}
where $\#$ denotes the pushforward measure. 

Enforcing Eq.~\ref{eq:fp} directly as a global distributional constraint is intractable. Instead, we minimize the local correction residual. Although the teacher suffers from severe truncation error when generating in a single step from pure noise, its local refinement from intermediate noise levels remains faithful to its original training objective. Evaluating the teacher on the student's partially re-masked draft therefore provides a valid descent direction toward the data manifold.

As shown in Fig.~\ref{fig:method}, we construct an end-to-end training pipeline. For each prompt $c$, we instantiate a nearly fully masked initialization $z_{\text{init}}$ at a high mask ratio $r_{\text{init}}$, with the unmasked positions filled by random codebook entries. The student generates a full-sequence draft in one shot:
\begin{equation}
\ell_\theta=g_\theta(z_{\text{init}},c),\qquad \hat{z}\sim p_\theta(\cdot\mid z_{\text{init}},c).
\end{equation}
To compute the local correction signal, a single teacher refinement is performed at a randomly sampled intermediate mask ratio $r \sim \pi$:
\begin{equation}
\hat{z}_T=\mathcal{T}_\phi^{r}\big(\mathcal{M}_r(\hat{z}),\,c\big). 
\end{equation}
The pair $(\hat{z},\hat{z}_T)$ constitutes the empirical fixed-point residual. The training loss, detailed in Sec.~\ref{subsec:drift}, penalizes this discrepancy in a continuous feature space, pulling the student toward the teacher's local refinement and enforcing the fixed-point condition across masking levels via the sampled $r$. 

Unlike prior discrete distillation methods that require trajectory-level alignment or auxiliary learnable networks, our formulation optimizes a single fixed-point objective via local corrections derived from the student's own drafts. Moreover, since both training and inference start from the same masked initialization $z_{\text{init}}$, the train- and test-time operating points coincide by construction.

\subsection{Lifted Particle Optimization}
\label{subsec:drift}

To compute the fixed-point residual in a meaningful metric space, we lift the discrete tokens into a continuous feature space. Because the native output of a discrete diffusion model is a sequence of categorical tokens $\hat{z}\in\mathcal{V}^L$, no well-defined distance exists between token sequences. We bridge this gap by sequentially passing tokens through the codebook embedding $E$, the decoder $\mathcal{D}$, and a frozen feature backbone $\Phi$. For a student draft $\hat{z}_i$ and its corresponding teacher refinement $\hat{z}_{T,i}$, the lifted spatial features are 
$X_i=\Phi\big(\mathcal{D}(E[\hat{z}_i];\,c_i)\big)$ and $Y_i=\Phi\big(\mathcal{D}(E[\hat{z}_{T,i}];\,c_i)\big)$.

In this continuous space, following~\cite{deng2026generative}, we employ a multi-bandwidth drift objective to construct the optimization signal from these paired features. Within a mini-batch, the empirical drift vector $V_h(X_i^f)$ at spatial position $f$ and kernel bandwidth $h$ is defined by an attractive force toward the teacher targets and a repulsive force from in-batch negative samples:
\begin{equation}
V_h(X_i^f)=\sum_{j}a_{ij}^{+}(h)\,(Y_j^f-X_i^f)\;-\;\sum_{j}a_{ij}^{-}(h)\,\big(X_{\sigma(j)}^f-X_i^f\big).
\end{equation}
Here, $\{X_{\sigma(j)}\}$ are cyclic in-batch shifts representing the student marginal, and $a^{\pm} \propto \exp(-\|X_i^f-\cdot\|/h)$ are doubly-normalized softmax affinities derived from a Laplace kernel. The student is optimized to match the displaced target via a stop-gradient regression:
\begin{equation}\label{eq:drift}
\mathcal{L}_{\mathrm{drift}}=\sum_{h\in\mathcal{H}}\frac{1}{Z_h}\,\mathbb{E}_{i,f}\,\big\|X_i^f-\mathrm{sg}\big(X_i^f+V_h(X_i^f)\big)\big\|^2,
\end{equation}
where $\mathcal{H}$ is a set of bandwidths and $Z_h$ is an RMS normalizer that equalizes gradient scales across bandwidths.

\subsection{Differentiable Routing via the Discrete Bottleneck}
\label{subsec:stein}

The drift objective operates in continuous space, yet the student outputs discrete tokens $\hat{z}\sim p_\theta$. Standard continuous relaxations, such as probability-weighted soft embeddings $\tilde{e}_i=\sum_v p_{\theta,iv}\,E_v$, produce off-manifold vectors. Feeding such off-manifold inputs into the teacher network elicits out-of-distribution refinements and systematically biases the fixed-point target $\hat{z}_T$.

To maintain differentiability while preserving manifold consistency, we employ a hard-forward, soft-backward straight-through estimator (STE)~\cite{bengio2013estimating}:
\begin{equation}\label{eq:stein}
e_i^{\text{STE}} = E[\hat z_i] + \tilde e_i - \mathrm{sg}(\tilde e_i).
\end{equation}
In the forward pass, $e_i^{\text{STE}} \equiv E[\hat{z}_i]$, ensuring that both the decoder and the teacher network evaluate only valid codebook embeddings and preserving the integrity of the local correction signal. In the backward pass, gradients route through the soft embedding, yielding $\partial e_i^{\text{STE}} / \partial \ell_{i,v} = \partial \tilde{e}_i / \partial \ell_{i,v}$, which translates the continuous drift force from the feature space back to the student logits.

This differentiable pixel-to-logit pathway natively supports image-level adversarial supervision. We optionally augment the training with an unconditional GAN objective $\mathcal{L}_{\text{gan}}$ applied to the decoded student images. Because the drift loss already enforces conditional and structural correspondence with the teacher targets, the unconditional discriminator serves solely to enhance high-frequency details and perceptual realism. The final training objective is:
$$\mathcal{L}_{\text{total}} = \mathcal{L}_{\text{drift}} + \lambda \mathcal{L}_{\text{gan}},$$
where $\lambda$ is a hyperparameter balancing the adversarial penalty against the fixed-point correction.

\section{Experiment}
\label{sec:exp}

We evaluate Fixed-Point Distillation on both class-conditional and
text-to-image generation, addressing three questions:
1)~Can a single-step discrete generator match multi-step teacher quality
without an auxiliary score network?
2)~How critical are the straight-through estimator and the student-driven
refinement source to the fixed-point formulation?
3)~How should the lifted drift space be configured in terms of feature
layers, spatial granularity, and kernel bandwidths?

\subsection{Experimental Settings}

\noindent
\textbf{Datasets and Metrics.} We evaluate on both class-conditional and text-to-image generation. For class-conditional generation, we use ImageNet~\cite{deng2009imagenet}; for text-to-image generation, we use a 400K-image subset of LAION~\cite{schuhmann2022laion}. Note that the drift loss requires only conditioning signals, i.e., class labels or text captions, to query the frozen teacher, while real images enter the pipeline solely through the unconditional discriminator. For class-conditional evaluation, we generate 50,000 images and report Fréchet Inception Distance~\cite{heusel2017gans}, Inception Score~\cite{salimans2016improved}, and Precision and Recall~\cite{kynkaanniemi2019improved}. For text-to-image evaluation, we report results on the GenEval~\cite{ghosh2023geneval} benchmark.

\noindent
\textbf{Implementation Details.} We initialize the student from pre-trained MaskGIT and MaskGen-L for class-conditional and text-to-image generation, respectively. The frozen feature backbone is DINOv3 ViT-B/16~\cite{simeoni2025dinov3}, from which we extract multi-scale spatial features at blocks $\{2, 5, 8, 11\}$ with drift bandwidths $\mathcal{H}=\{0.02, 0.05, 0.2\}$. Both the feature backbone and the VQ decoder remain frozen throughout training. All models are optimized with AdamW at a learning rate of $1\times10^{-5}$, using a batch size of 32 for MaskGIT and 64 for MaskGen-L.

We optionally incorporate a StyleGAN-XL~\cite{sauer2022stylegan} discriminator to enhance perceptual quality. The discriminator is unconditional and trained on real images only, ensuring that all conditional generation capability originates from the drift loss rather than from adversarial supervision.

\noindent \textbf{Baselines.} For class-conditional generation, we compare against the MaskGIT~\cite{besnier2023pytorch} teacher at varying inference steps, as well as recent discrete distillation methods: SDTT~\cite{deschenaux2024beyond}, di4c~\cite{hayakawa2024distillation}, ReDi~\cite{yoo2025redi}, and DiMO~\cite{zhu2025di}. For text-to-image generation, we compare against the MaskGen teacher, the discrete baseline DiMO~\cite{zhu2025di}, and a broad range of continuous distillation methods including InstaFlow~\cite{liu2023instaflow}, SiD-LSG~\cite{zhou2024long}, LCM~\cite{luo2023latent}, TDM~\cite{luo2025learning}, ADD~\cite{sauer2024adversarial}, and DMD2~\cite{yin2024improved}. 

\begin{table}[t!]
\setlength{\tabcolsep}{2.5pt}
\centering
\caption{
Comparison of various text-to-image generation models on the GenEval benchmark.
The symbols $\uparrow$ and $\downarrow$ indicate that higher and lower scores are preferable, respectively.
``\#Model Params'' is the inference parameter count.
}
\small
\label{tab:main_t2i}
\begin{tabular}{lccccccccccc}
\toprule[1pt]
\multirow{2}{*}{Methods} & \multirow{2}{*}{Steps\,$\downarrow$} & \multirow{2}{*}{\shortstack{\#Model\\Params}} & \multirow{2}{*}{\shortstack{Aux.\,Score\\Net}} & \multicolumn{7}{c}{GenEval~$\uparrow$} \\
\cmidrule(lr){5-11}
 &  &  &  & Single & Two & Count. & Colors & Pos. & Color Attr. & Overall \\
\midrule[0.8pt]
\multicolumn{11}{l}{\textit{\color{gray}Foundation Models}} \\[2pt]
LDM~\cite{rombach2022high}             & 50 & 1.4B & -     & 0.92 & 0.29 & 0.23 & 0.70 & 0.02 & 0.05 & 0.37 \\
DALLE~2~\cite{ramesh2022hierarchical}  & -  & 4.2B & -     & 0.94 & 0.66 & 0.49 & 0.77 & 0.10 & 0.19 & 0.52 \\
SDXL~\cite{podell2023sdxl}             & 50 & 2.6B & -     & 0.98 & 0.74 & 0.39 & 0.85 & 0.15 & 0.23 & 0.55 \\
MaskGen~\cite{kim2025democratizing}    & 16 & 0.6B & -     & 0.97 & 0.55 & 0.38 & 0.80 & 0.08 & 0.14 & 0.48 \\
\midrule
\multicolumn{11}{l}{\textit{\color{gray}Continuous Distillation}} \\[2pt]
InstaFlow~\cite{liu2023instaflow}              & 1 & 0.9B & \xmark & 0.88 & 0.21 & 0.20 & 0.66 & 0.03 & 0.03 & 0.33 \\
SiD-LSG~\cite{zhou2024long}                    & 1 & 0.9B & \cmark & 0.93 & 0.37 & 0.21 & 0.57 & 0.03 & 0.03 & 0.36 \\
RG-LCM (HPS)~\cite{li2024reward}               & 2 & 0.9B & \xmark & 0.97 & 0.54 & 0.35 & 0.82 & 0.07 & 0.14 & 0.48 \\
TDM~\cite{luo2025learning}                     & 4 & 0.9B & \cmark & 0.99 & 0.57 & 0.49 & 0.78 & 0.09 & 0.09 & 0.50 \\
SDXL-LCM~\cite{luo2023latent}                  & 1 & 2.6B & \xmark & 0.75 & 0.11 & 0.14 & 0.59 & 0.01 & 0.03 & 0.27 \\
SDXL-LCM~\cite{luo2023latent}                  & 4 & 2.6B & \xmark & 0.99 & 0.57 & 0.39 & 0.86 & 0.09 & 0.18 & 0.51 \\
SDXL-Turbo~\cite{sauer2024adversarial}         & 1 & 2.6B & \xmark & 0.99 & 0.65 & 0.52 & 0.87 & 0.12 & 0.19 & 0.55 \\
SDXL-DMD2~\cite{yin2024improved}               & 1 & 2.6B & \cmark & 0.99 & 0.68 & 0.48 & 0.90 & 0.08 & 0.19 & 0.55 \\
\midrule
\multicolumn{11}{l}{\textit{\color{gray}Discrete Distillation}} \\[2pt]
MaskGen-DiMO~\cite{zhu2025di}                  & 1 & 0.6B & \cmark & 0.93 & 0.39 & 0.35 & 0.74 & 0.07 & 0.08 & 0.42 \\
\rowcolor{gray!20}
MaskGen-FPD (Ours)                             & 1 & 0.6B & \xmark & 0.96 & 0.36 & 0.43 & 0.76 & 0.09 & 0.09 & 0.45 \\
\bottomrule
\end{tabular}
\end{table}

\begin{table}[t!]
\centering
\caption{
Quantitative results on class-conditional ImageNet-256 with MaskGit teacher.
}
\small
\label{tab:main_c2i}
\begin{tabular}{lcccccc}
\toprule[1pt]
Method & Steps $\downarrow$ & \shortstack{Aux.\,Score\\Net} & FID $\downarrow$ & IS $\uparrow$ & Prec. $\uparrow$ & Rec. $\uparrow$ \\
\midrule[0.8pt]
\multicolumn{7}{l}{\textit{\color{gray}Teacher}} \\[2pt]
MaskGit~\cite{besnier2023pytorch} & 16 & -     & 6.60  & 224 & 0.83 & 0.40 \\
MaskGit~\cite{besnier2023pytorch} & 8  & -     & 6.66  & 222 & 0.83 & 0.40 \\
MaskGit~\cite{besnier2023pytorch} & 4  & -     & 10.73 & 192 & 0.75 & 0.31 \\
MaskGit~\cite{besnier2023pytorch} & 2  & -     & 91.35 & 13  & 0.18 & 0.16 \\
\midrule
\multicolumn{7}{l}{\textit{\color{gray}Discrete Distillation}} \\[2pt]
SDTT~\cite{deschenaux2024beyond} & 4 & \xmark & 8.97  & 205 & 0.88 & 0.41 \\
SDTT~\cite{deschenaux2024beyond} & 1 & \xmark & 90.40 & 14  & 0.31 & 0.13 \\
di4c~\cite{hayakawa2024distillation}      & 4 & \xmark & 6.79 & 209 & -    & -    \\
di4c-d~\cite{hayakawa2024distillation}    & 4 & \xmark & 6.57 & 214 & -    & -    \\
ReDi$^1$~\cite{yoo2025redi}               & 4 & \xmark & 7.58 & 228 & 0.87 & 0.46 \\
ReDi$^2$~\cite{yoo2025redi}               & 4 & \xmark & 7.86 & 240 & 0.87 & 0.44 \\
ReDi$^3$-distill~\cite{yoo2025redi}       & 1 & \xmark & 11.68 & 182 & 0.83 & 0.44 \\
DiMO~\cite{zhu2025di}                     & 1 & \cmark & 6.91 & 214 & 0.83 & 0.38 \\
\rowcolor{gray!20}
FPD (Ours)                                & 1 & \xmark & 6.90 & 215 & 0.81 & 0.35 \\
\bottomrule
\end{tabular}
\end{table}

\subsection{Main Results}

We compare our method against existing approaches on text-to-image and class-conditional generation in Tab.~\ref{tab:main_t2i} and Tab.~\ref{tab:main_c2i}, respectively. Both tables additionally report whether each distillation method requires an auxiliary score network, an online learnable network that approximates the score of the student's output distribution and is updated jointly with the student. Such a network is typically the same size as the teacher, effectively doubling training compute.

\begin{figure}[t!]
    \centering
    \includegraphics[width=1.0\linewidth]{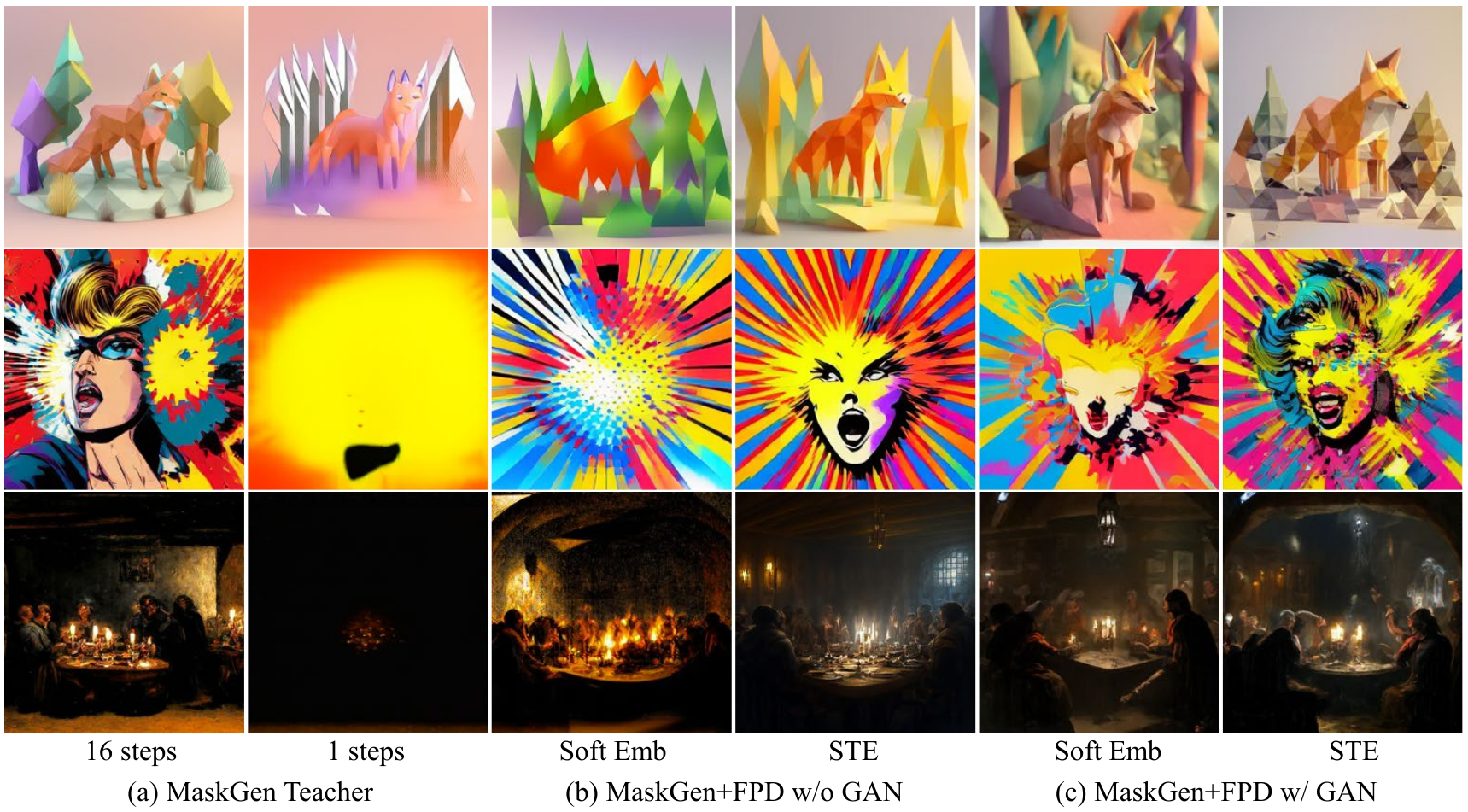}
    \caption{Training strategy comparisons. We compare the multi-step teacher baselines with our single-step FPD variants, demonstrating the superiority of the Straight-Through Estimator in preserving structural integrity, and the perceptual enhancements introduced by the auxiliary GAN loss.}
    \label{fig:abla_gan_and_soft}
\end{figure}

\noindent
\textbf{Text-to-Image Generation.}
Tab.~\ref{tab:main_t2i} demonstrates the efficacy of our approach in achieving high-quality single-step generation without the severe degradation typical of extreme distillation. On the GenEval text-to-image benchmark, the distilled MaskGen-FPD achieves an overall score of 0.45 in a single step using an efficient 0.6B parameter budget. This outperforms the competing one-step discrete baseline MaskGen-DiMO, which scores 0.42. Notably, our method yields significant gains in the Counting metric, reaching 0.43 compared to DiMO's 0.35. Furthermore, it successfully retains roughly 94\% of the overall performance of the 16-step MaskGen teacher, which scores 0.48. Fig.~\ref{fig:teaser} and Fig.~\ref{fig:abla_gan_and_soft} visually
demonstrate the effectiveness of our method, whereas one-step
generation directly from the teacher collapses entirely.

\noindent
\textbf{Class-Conditional Generation.}
This robustness extends to class-conditional ImageNet-256 generation, as detailed in Tab.~\ref{tab:main_c2i}. Our method establishes a highly competitive one-step FID of 6.90 and an Inception Score of 215. While prior discrete distillation approaches fail severely at a single step, such as SDTT degrading to an FID of 90.40, or struggle to close the quality gap, as seen with ReDi$^3$-distill at 11.68, our single-step output rivals the quality of multi-step samplers. Specifically, it eclipses the 4-step ReDi$^1$, which yields an FID of 7.58, and performs on par with the recent one-step DiMO baseline at 6.91, while bridging the performance gap toward the 16-step MaskGit teacher, which holds an FID of 6.60. These empirical results confirm that the fixed-point formulation successfully condenses the iterative generation process into a single evaluation while strictly preserving visual fidelity and structural alignment.

\begin{table}[t]
\centering
\caption{Ablations on gradient routing and teacher refinement, evaluated on GenEval.
(a)~Impact of the discrete bottleneck strategy and the optional adversarial loss.
\emph{Soft Emb.} computes a probability-weighted codebook mixture;
\emph{STE} uses hard-sampled tokens in the forward pass.
(b)~Refinement source for the teacher target: \emph{Student} re-masks the student's
own draft; \emph{Random} initialises all positions randomly.}
\label{tab:architecture_features}
\setlength{\tabcolsep}{7pt}
\renewcommand{\arraystretch}{1.08}
\begin{subtable}[t]{0.48\linewidth}
\centering
\caption{Gradient Routing and Adversarial Loss}
\label{tab:ablation_path_gan}
\begin{tabular}{lcc}
\toprule
Estimator & w/ $\mathcal{L}_\text{gan}$ & Overall $\uparrow$ \\
\midrule
Soft Emb. &        & 0.38 \\
Soft Emb. & \cmark & 0.42 \\
STE       &        & 0.43 \\
STE       & \cmark & \textbf{0.45} \\
\bottomrule
\end{tabular}
\end{subtable}
\hfill
\begin{subtable}[t]{0.48\linewidth}
\centering
\caption{Refinement Source}
\label{tab:ablation_teacher_refinement}
\begin{tabular}{lc}
\toprule
Draft Source & Overall $\uparrow$ \\
\midrule
Random  & 0.22 \\
Student & \textbf{0.43} \\
\bottomrule
\end{tabular}
\end{subtable}
\end{table}

\begin{table}[t]
\centering
\caption{Feature space ablations on GenEval.
(a)~Representation space and kernel bandwidths for the drift objective.
(b)~Layer selection and spatial granularity of the frozen feature backbone
(DINOv3-B, 12 layers). \emph{Patch Grid} indicates whether features are
kept on a $4{\times}4$ spatial grid or globally average-pooled into a single
vector.}
\label{tab:mechanism_ablation}
\setlength{\tabcolsep}{7pt}
\renewcommand{\arraystretch}{1.08}
\begin{subtable}[t]{0.48\linewidth}
\centering
\caption{Drift Space and Kernel Bandwidths}
\label{tab:ablation_drift_space_scale}
\begin{tabular}{lcc}
\toprule
Space & Bandwidths & Overall $\uparrow$ \\
\midrule
Pixel   & \{0.02, 0.05, 0.2\} & 0.21 \\
Feature & \{0.05\}             & 0.41 \\
Feature & \{0.02, 0.05, 0.2\} & \textbf{0.43} \\
\bottomrule
\end{tabular}
\end{subtable}
\hfill
\begin{subtable}[t]{0.48\linewidth}
\centering
\caption{Backbone Layer Selection}
\label{tab:ablation_dino_groups}
\begin{tabular}{lcc}
\toprule
Blocks & Patch Grid & Overall $\uparrow$ \\
\midrule
\{2,5,8,11\} &        & 0.34 \\
\{2,5\}      & \cmark & 0.35 \\
\{8,11\}     & \cmark & 0.42 \\
\{2,5,8,11\} & \cmark & \textbf{0.43} \\
\bottomrule
\end{tabular}
\end{subtable}
\end{table}

\subsection{Ablation Studies}

\noindent \textbf{Gradient Routing and Adversarial Supervision.}
Tab.~\ref{tab:ablation_path_gan} and Fig.~\ref{fig:abla_gan_and_soft} evaluate the
gradient routing strategy alongside the optional adversarial loss. Replacing
the soft embedding path with the Straight-Through Estimator improves the
GenEval score from 0.38 to 0.43, confirming that preserving exact hard
sampling during the forward pass prevents out-of-distribution teacher
evaluations and maintains train--test manifold consistency. As illustrated
in Fig.~\ref{fig:abla_gan_and_soft}, soft embeddings introduce visible
artifacts and structural distortions, since the teacher and decoder receive
off-manifold inputs that deviate from their training distribution. This
result also shows that the fixed-point framework is effective on its own,
establishing a strong generative baseline without any adversarial
supervision. Adding $\mathcal{L}_\text{gan}$ further improves both routing
paths, with the STE combination reaching the highest score of 0.45. The
unconditional discriminator primarily enhances high-frequency details and
perceptual sharpness, acting as a complementary booster rather than the
primary source of semantic alignment.

\noindent \textbf{Student-Driven Teacher Refinement.}
Tab.~\ref{tab:ablation_teacher_refinement} validates the fixed-point formulation
by comparing the refinement source. When the teacher refines a random token
sequence instead of the student's own draft, performance collapses from 0.43
to 0.22. This confirms that the teacher's value lies in providing localized,
state-dependent corrections to the student's current output rather than acting
as a standalone generator. Intuitively, the student's draft already captures
coarse structure from its conditioning signal; the teacher then sharpens this
structure through targeted refinement of the re-masked positions. When
this coupling is removed, the teacher must reconstruct the entire sequence
from scratch, and the resulting drift vectors bear no relation to the
student's generative state, leaving the optimization without a coherent
descent direction.

\noindent \textbf{Lifted Drift Space.}
Tab.~\ref{tab:ablation_drift_space_scale} and Tab.~\ref{tab:ablation_dino_groups}
validate the configuration of the drift objective. Computing the loss in raw
pixel space yields only 0.21, as pixel-level distances are dominated by
low-level variations and fail to capture the semantic structure needed to
guide distillation. Lifting tokens into a pre-trained feature space is
therefore essential. Within that space, the choice of layers and spatial
granularity both matter. Relying on global average pooling discards spatial
correspondence between the student and teacher outputs, while using only
shallow layers provides features that lack sufficient semantic
abstraction. A full multi-layer spatial configuration over blocks
$\{2,5,8,11\}$ with a $4{\times}4$ patch grid captures both mid-level
structure and high-level semantics, reaching 0.43. Finally, a multi-scale
bandwidth set $\mathcal{H}$ outperforms a single bandwidth by allowing the
drift kernel to simultaneously enforce coarse global alignment at large $h$
and fine-grained local discrimination at small $h$.

\section{Conclusion}
\label{sec:conclusion}

We presented Fixed-Point Distillation, a framework enabling high-fidelity single-step generation for discrete diffusion models. By formulating distillation as a fixed-point matching problem, we lift discrete tokens into a continuous feature space, allowing for efficient particle optimization via a drift loss without auxiliary score networks. To resolve the discrete bottleneck, we employed a straight-through estimator that ensures manifold consistency for the teacher evaluations while routing structured continuous gradients back to the student logits. This differentiable pathway additionally supports unconditional adversarial supervision to enhance perceptual realism. Experiments on class- and text-condition generation benchmarks achieve competitive results compared to existing methods, condensing the iterative sampling process into a single evaluation while preserving structural alignment and visual quality.

{\small
\bibliographystyle{ieee_fullname}
\bibliography{refbib}
}

\end{document}